\documentclass{article} % For LaTeX2e
\usepackage{nips15submit_e,times}
\usepackage{url}
\usepackage{graphicx} % more modern
\usepackage{amsthm}
\usepackage{amsmath}
\usepackage{amsfonts}
\usepackage{bbm}
\usepackage{titlesec}
\usepackage{caption}
\usepackage{wrapfig}
\usepackage[toc,page]{appendix}

\usepackage[tight,footnotesize]{subfigure}

\title{Phenotyping of Clinical Time Series with LSTM Recurrent Neural Networks}

\author{
Zachary C. Lipton  \\ %\thanks{Author website: http://zacklipton.com } \\
Computer Science \& Engineering\\
UC San Diego\\
La Jolla, CA 92093, USA \\
\texttt{zlipton@cs.ucsd.edu} \\
\And
David C. Kale \\ %\thanks{Author website: http://www-scf.usc.edu/~dkale/ }\\
Computer Science \\
USC \\ %University of Southern California \\
Los Angeles, CA 90089 \\
\texttt{dkale@usc.edu} \\
 \And
 Randall C. Wetzel \\
Whittier Virtual PICU\\
Children's Hospital LA \\
 Los Angeles, CA 90027 \\
 \texttt{rwetzel@chla.usc.edu} \\
}

\nipsfinalcopy % Uncomment for camera-ready version

\begin{document}

\maketitle

\begin{abstract}
We present a novel application of LSTM recurrent neural networks 
to multilabel classification of diagnoses given variable-length  time series of clinical measurements. 
Our method outperforms a strong baseline on a variety of metrics.

%and convulutional nets, 
% predicting $429$ diagnoses from just $13$ frequently but irregularly sampled features, achieving AUC of $.89$ and F1 of $.238$. 
% \todo{Update once we've finished draft.}
\end{abstract}

\section{Introduction}
Recurrent neural networks (RNNs),
in particular those based on Long Short-Term Memory (LSTM) \cite{hochreiter1997long}, 
powerfully model varying-length sequential data,
achieving state-of-the-art results for problems 
spanning natural language processing, image captioning, handwriting recognition,
and genomic analysis \cite{auli2013joint, sutskever2014sequence, vinyals2014show, karpathy2014deep, liwicki2007novel, graves2009novel, pollastri2002improving, vohradsky2001neural, xu2007inference}. 
LSTM RNNs can capture long range dependencies and nonlinear dynamics. 
Clinical time series data,
as recorded in the pediatric intensive care unit (PICU), 
exhibit these properties and others, including irregular sampling and non-random missing values \cite{marlin:ihi2012}. 
Symptoms of acute respiratory distress syndrome, for example, often do not appear for $24$-$48$ hours after lung injury \cite{mason2010murray}. 
Other approaches like Markov models, conditional random fields, and Kalman filters deal with sequential data, 
but are ill-equipped to learn long-range dependencies.
Some models require domain knowledge or feature engineering, 
offering less chance for serendipitous discovery. 
Neural networks learn representations, 
potentially discovering unforeseen structure.

This paper presents a preliminary empirical study 
of LSTM RNNs applied to \textit{supervised phenotyping} of multivariate PICU time series.
We classify each episode as having one or more diagnoses 
from among over one hundred possibilities.
Prior works have applied RNNs to health data, 
including electrocardiograms \cite{silipo1998artificial, amari1998adaptive, ubeyli2009combining} and glucose measurements \cite{tresp:nips1997}. 
RNNs have also been used for prediction problems in genomics \cite{pollastri2002improving, xu2007inference, vohradsky2001neural}. A variety of works have applied feed-forward neural networks to health data for prediction and pattern mining, but none have used RNNs or directly handled variable length sequences
\cite{dabek2015neural,rughani2010use,lasko:plosone2013,che:kdd2015}.
%predict psychological diagnoses and survival in traumatic brain injury, respectively, with feed-forward neural networks but neither use RNNs.
To our knowledge, this work is the first 
to apply modern LSTMs to a large data set 
of multivariate clinical time series.
Our experiments show that LSTMs can successfully classify clinical time series from raw measurements, 
naturally handling challenges like variable sequence length 
and high dimensional output spaces.

\section{Data Description}
Our experiments use a collection of fully anonymized clinical time series 
extracted from the electronic health records system at Children's Hospital LA \cite{marlin:ihi2012,che:kdd2015} as a part of an IRB-approved study. 
The data consist of $10,401$ PICU episodes, 
each a multivariate time series of 13 variables 
including vital signs, lab results, and subjective assessments.
The episodes vary in length from 12 hours to 30 days.
Each episode has zero or more diagnostic labels 
from an in-house taxonomy, similar to ICD-9 codes, 
used for research and billing. 
There are 128 distinct labels 
indicating a variety of conditions, 
such as acute respiratory distress, congestive heart failure, seizures, renal failure, and sepsis.

The original data are irregularly sampled multivariate time series with missing values and occasionally missing variables.
% For example, an episode will have no end-tidal CO$_2$ measurements if the patient is not intubated).
We resample all time series to an hourly rate (similar to \cite{marlin:ihi2012}), taking the mean measurement within each one hour window and filling gaps by propagating measurements forward or backward. When time series are missing entirely, we impute a clinically normal value.\footnote{Many variables are recorded at rates proportional to how quickly they change, and when a variable is entirely absent, it is often because clinical staff believed it to be normal and chose not to measure it.} 
We rescale variables to a $[0,1]$ interval using ranges defined by clinical experts.
% (dependent upon age and gender in the case of vitals, such as heart rate).

\section{Methods and Experiments}
We cast the problem of phenotyping clinical time series as multilabel classification,
and our proposed LSTM RNN
uses memory cells %\ref{fig:figure1} 
with forget gates as described in \cite{gers2000learning} 
but without peephole connections as described in \cite{gers2003learning}. 
As output, we use a fully connected layer atop the highest LSTM layer, with a sigmoid activation function 
because the problem is multilabel.
Binary cross-entropy is the loss at each output node.
Among architectures that we tested,
the simplest and most effective passes over the data in chronological order, outputting predictions 
only at the final sequence step.

\begin{wrapfigure}{r}{0.35\textwidth}
  \begin{center}
	\vspace{-20pt}
	\includegraphics[width=0.4\textwidth]{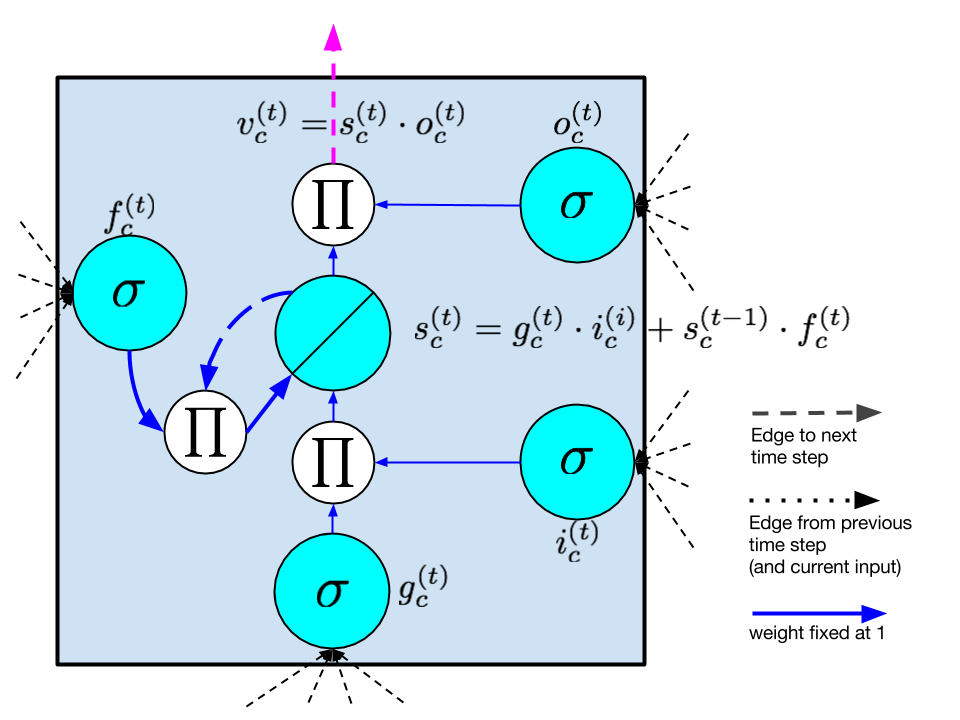}
    \vspace{-25pt}
  \end{center}
  \caption{LSTM memory cell with \\ forget gate as depicted in \cite{lipton2015critical}}
\end{wrapfigure} 
We train the network using stochastic gradient descent with momentum. 
Absent momentum, the variance of the gradient is large, 
and single examples occasionally destroy the model.
Interestingly, the presence of exploding gradients had no apparent connection to the loss on the particular example that caused it. 
To combat exploding gradients, 
we experimented with $\ell_2^2$ weight decay, gradient clipping, 
and truncated back-propagation. We test various settings for the number of layers and nodes, choosing the best using validation performance.
% Ultimately, with $128$ hidden nodes and a maximum sequence length under $200$ further techniques were not necessary. 

We evaluate the LSTM-based models 
against a variety of baselines. 
All models are trained on $80\%$ of the data 
and tested on $10\%$. 
The remaining $10\%$ is used for hyper-parameter optimization, e.g., regularization strength and early stopping. 
We report micro- and macro-averaged area under the ROC curve (AUC) and F1 score.
We also report \textit{precision at $10$}, 
which captures the fraction of true diagnoses 
among the model's top 10 predictions, with a best possible score of $0.2818$ on these data.  
We provide results for a \textit{base rate} model that predicts diagnoses in descending order by incidence to provide a minimum performance baseline. %which aids interpretation of , 
%given varying class imbalance among labels. 
Logistic regression with $\ell_2^2$ regularization and 
carefully engineered features encoding domain knowledge formed a strong baseline.
% likely due to the small number of variables in our time series 
% and our preprocessing procedure. 
Nonetheless, an LSTM with two layers of 128 hidden units achieved the best overall performance on all metrics but precision at 10, while using only raw time series as input.

\begin{figure*}[hbt!]
\centering
\footnotesize
% \scriptsize
\begin{tabular}{|l|l|l|l|l|l|}
 \hline
 \multicolumn{6}{|c|}{Overall classification performance for 128 PICU phenotypes} \\
 \hline
  Model & Micro AUC & Macro AUC & Micro F1 & Macro F1 & Precision at 10 \\
 \hline
Base rate & 0.7170 & 0.5 & 0.1366 & 0.0339 & 0.0753 \\ % 0.2281
Linear, last $12$ hours raw data & 0.8041 & 0.7286 & 0.2263 & 0.1004 & 0.0986 \\
Linear, engineered features & 0.8277 & 0.7628 & 0.2498 & 0.1254 & \textbf{0.1085} \\
% LSTM(2,128), trained on 429 &
LSTM, 2 layers, 128 nodes each &
\textbf{0.8324} & \textbf{0.7717} & \textbf{0.2577} & \textbf{0.1304} & 0.1078 \\
\hline
\end{tabular} \hspace{-0.2in}
\vspace{-0.2in}
\label{fig:experimental-results}
\end{figure*}

\section{Discussion}
Our results indicate that LSTM RNNs can be successfully applied 
to the problem of phenotyping critical care patients given clinical time series data.
Promising early experiments with gradient normalization suggest that we can improve our results further.
Our next steps to advance this research
include advanced optimization and regularization strategies,
techniques to directly handle missing values and irregular sampling, and extending this work to a larger PICU data set
with a richer set of measurements, 
including treatments and medications. 
Additionally, there remain many questions about the interpretability of neural networks when applied to complex medical problems. 
We are developing methods to expose the patterns of health and illness learned by LSTMs to clinical users and to make practical use of the distributed representations 
learned by LSTMs in applications like patient similarity search.

\section*{Acknowledgments}
Zachary C. Lipton was supported by the Division of Biomedical Informatics at the University of California, San Diego, via training grant (T15LM011271) from the NIH/NLM.
David Kale was supported by the Alfred E. Mann Innovation in Engineering Doctoral Fellowship.
% The authors would like to thank Randall Wetzel of Children's Hospital Los Angeles for his assistance in understanding and analyzing clinical data. 
The VPICU was supported by grants from the Laura P. and Leland K. Whittier Foundation.

We acknowledge NVIDIA Corporation for Tesla K40 GPU hardware donation and Professors Charles Elkan and Julian McAuley for their support.

\newpage

\bibliography{lstm-icu}
\bibliographystyle{unsrt}

\newpage

\end{document}